\documentclass[11pt,a4paper]{article}
\usepackage[]{authblk}
\usepackage[hyperref]{acl2021}
\usepackage{times}
\usepackage{latexsym}

\usepackage{microtype}

\aclfinalcopy 

\usepackage{todonotes}
\usepackage{graphicx}
\usepackage{amsmath}
\usepackage{cleveref}

\newcommand{\argmax}{\operatornamewithlimits{arg\,max}}
\usepackage{multirow}
\usepackage{xcolor}
\usepackage{tabularx}
\newcolumntype{L}{>{\raggedright\arraybackslash}X}    
\usepackage{subcaption}
\usepackage{times}

\newsavebox\CBox
\def\mathcellbold#1{\sbox\CBox{#1}\resizebox{\wd\CBox}{\ht\CBox}{\boldmath{#1}}}

\title{Investigation on Data Adaptation Techniques \\ for Neural Named Entity Recognition}
\author[ ]{Evgeniia Tokarchuk\thanks{\hspace{0.5em}Work completed while studying at RWTH Aachen University.}}
\author[$\dagger$]{David Thulke}
\author[$\dagger$]{Weiyue Wang}
\author[$\dagger$]{Christian Dugast}
\author[$\dagger$]{Hermann Ney}
\affil[$\ast$]{Informatics Institute, University of Amsterdam}
\affil[$\dagger$]{Human Language Technology and Pattern Recognition Group \authorcr Computer Science Department \authorcr RWTH Aachen University}
\affil[ ]{\texttt{e.tokarchuk@uva.nl}}
\affil[ ]{\texttt{\{thulke,wwang,dugast,ney\}@cs.rwth-aachen.de}}
\date{}

\begin{document}
\maketitle


\begin{abstract}
Data processing is an important step in various natural language processing tasks. As the commonly used datasets in named entity recognition contain only a limited number of samples, it is important to obtain additional labeled data in an efficient and reliable manner. A common practice is to utilize large monolingual unlabeled corpora. Another popular technique is to create synthetic data from the original labeled data (data augmentation). In this work, we investigate the impact of these two methods on the performance of three different named entity recognition tasks.
\end{abstract}

\section{Introduction}
Recently, deep neural network models have emerged in various fields of natural language processing (NLP) and replaced the mainstream position of conventional count-based methods \citep{LB16,VS17,SS16}. In addition to providing significant performance improvements, neural models often require high hardware conditions and a large amount of clean training data. However, there is usually only a limited amount of cleanly labeled data available, so techniques such as data augmentation and self-training are commonly used to generate additional synthetic data.

Significant progress has been made in recent years in designing data augmentations for computer vision (CV) \cite{KrizhevskySH12}, automatic speech recognition (ASR) \cite{ParkCZCZCL19}, natural language understanding (NLU) \cite{HouLCL18} and machine translation (MT) \cite{wang-etal-2018-switchout} in supervised settings. In addition, semi-supervised approaches using self-training techniques \cite{BlumM98} have shown promising performance in conventional named entity recognition (NER) systems \cite{KB05,daume-iii-2008-cross,tackstrom-2012-nudging}. In this work, the effectiveness of self-training and data augmentation techniques on neural NER architectures is explored.

To cover different data situations, we select three different datasets: The English CoNLL 2003 \citep{SD03} dataset, which is the benchmark on which almost all NER systems report results, it is very clean and the baseline models achieve an F1 score of around 92.6\%; The English W-NUT 2017 \cite{D17} dataset, which is generated by users and contains inconsistencies, baseline models get an F1 score of around 52.7\%; The GermEval 2014 \cite{B14} dataset, a fairly clean German dataset with baseline scores of around 86.3\%\footnote{From here on, for the sake of simplicity, we omit the annual information of the datasets.}. We observe that the baseline scores on clean datasets such as CoNLL and GermEval can hardly be improved by data adaptation techniques, while the performance on the W-NUT dataset, which is relatively small and inconsistent, can be significantly improved.

\section{Related Work}

\subsection{State-of-the-art Techniques in NER}

\newcite{CollobertWBKKK11} advance the use of neural networks (NN) for NER, who propose an architecture based on temporal convolutional neural networks (CNN) over the sequence of words. Since then, many articles have suggested improvements to this architecture. \newcite{HuangXY15} propose replacing the CNN encoder in \newcite{CollobertWBKKK11} with a bidirectional long short-term memory (LSTM) encoder, while \newcite{LB16} and \newcite{chiu-nichols-2016-named} introduce a hierarchy into the architecture by replacing artificially designed features with additional bidirectional LSTM or CNN encoders. In other related work, \newcite{MesnilHDB13} have pioneered the use of recurrent neural networks (RNN) to decode tags. 

Recently, various pre-trained word embedding techniques have offered further improvements over the strong baseline achieved by the neural architectures. \newcite{AB18} suggest using pre-trained character-level language models from which to extract hidden states at the start and end character positions of each word to embed any string in a sentence-level context. In addition, the embedding generated by unsupervised representation learning \cite{peters-etal-2018-deep,devlin-etal-2019-bert,abs-1907-11692, TG2020} has been used successfully for NER, as well as other NLP tasks. In this work, the strongest model for each task is used as the baseline model.

\subsection{Data Adaptation in NLP}

In NLP, generating synthetic data using forward or backward inference is a commonly used approach to increase the amount of training data. 
In strong MT systems, synthetic data that is generated by back-translation is often used as additional training data to improve translation quality \cite{sennrich-etal-2016-improving}. A similar approach using backward inference is also successfully used for end-to-end ASR \cite{HayashiWZTHAT18}. In addition, back-translation, as observed by \newcite{YuDLZ00L18}, can create various paraphrases while maintaining the semantics of the original sentences, resulting in significant performance improvements in question answering. 

In this work, synthetic annotations, which are generated by forward inference of a model that is trained on annotated data, are added to the training data. The method of generating synthetic data by forward inference is also called self-training in semi-supervised approaches. \newcite{KB05} use self-training and co-training to recognize and classify named entities in the news domain. \newcite{tackstrom-2012-nudging} uses self-training to adapt a multi-source direct transfer named entity recognizer to different target languages, ``relexicalizing'' the model with word cluster features. \newcite{clark-etal-2018-semi} propose cross-view training, a semi-supervised learning algorithm that improves the representation of a bidirectional LSTM sentence encoder using a mixture of labeled and unlabeled data. 

In addition to the promising pre-trained embedding that is successfully used for various NLP tasks, the masked language modeling (MLM) can also be used for data augmentation. \newcite{kobayashi-2018-contextual} and \newcite{WuLZHH19} propose to replace words with other words that are predicted using the language model at the corresponding position, which shows promising performance on text classification tasks. Recently, \newcite{kumar2020data} discussed the effectiveness of such different pre-trained transformer-based models for data augmentation on text classification tasks. And for neural MT, \newcite{gao-etal-2019-soft} suggest replacing randomly selected words in a sentence with a mixture of several related words based on a distribution representation. In this work, we explore the use of MLM-based contextual augmentation approaches for various NER tasks.

\section{Self-training}
\label{sec:approach-self-training}

Though, the amount of annotated training data is limited for many NLP tasks, additional unlabeled data is available in most situations.
Semi-supervised learning approaches make use of this additional data.
A common way to do this is self-training \cite{KB05, tackstrom-2012-nudging, clark-etal-2018-semi}.

At a high level, it consists of the following steps:
\begin{enumerate}
\item An initial model is trained using the labeled data.
\item This model is used to annotate the additional unlabeled data.
\item A subset of this data is selected and used in addition to the labeled data to retrain the model.
\end{enumerate}

For the performance of the method it is critical to find a heuristic to select a good subset of the automatically labeled data.
The selected data should not introduce too many errors, but at the same time they should be informative, i.e. they should be useful to improve the decision boundary of the final model.
One selection strategy \cite{Drugman2016} is to calculate a confidence measure for all unlabeled sentences and to randomly sample sentences above a certain threshold.

We consider two different confidence measures in this work.
The first, hereinafter referred to as \(c_1\), is the posterior probability of the tag sequence $y$ given the word sequence $x$:
\begin{equation}
    c_1(y, x) = p(y \mid x) = \frac{e^{s(x,y)}}{\sum_{y'} e^{s(x, y')}}
\end{equation}
whereby $s(x, y)$ is the unnormalized log score assigned by the model to the sequence, consisting of an emission model \(q^E_i\) and transition model \(q^T\):
\[
  s(x, y_1^T) = \sum_{i=1}^T q^E_i(y_i \mid x) + q^T(y_i \mid y_{i-1})
\]

For the second confidence measure, we take into account the normalized tag scores at each position.
To get a confidence score for the entire sequence, we take the minimum tag score of all positions. 
Thus, \(c_2\) is defined as follows:
\begin{equation}
    c_2(y, x) = \min_i \frac{q^E_i(y_i \mid x) + q^T(y_i \mid y_{i-1})}{\sum_{y'_i} q^E_i(y'_i \mid x) + q^T(y'_i \mid y_{i-1})} 
\end{equation}

\section{MLM-based Data Augmentation}
\label{section:mlm}

Instead of using additional unlabeled data, we apply MLM-based data augmentation specifically for NER by masking and replacing original text tokens while maintaining labels.

For each masked token $x_{i}$:
\begin{equation}
    \hat{x}_{i}=\argmax_w p(x_{i}=w|\tilde{\textbf{x}})
\end{equation}
where $\hat{x}_{i}$ is the predicted token, $w \in V$ is the token from the model vocabulary and $\tilde{\textbf{x}}$ is the original sentence with $x_i=\texttt{[MASK]}$.

There are several configurations that can affect the performance of the data augmentation method: Techniques of selecting the tokens to be replaced, the order of token replacement in case of multiple replacement and the criterion for selecting the best tokens from the predicted ones. This section studies the effect of these configurations.

\subsection{Sampling}
\begin{figure*}
    \centering
    \includegraphics[width=\linewidth]{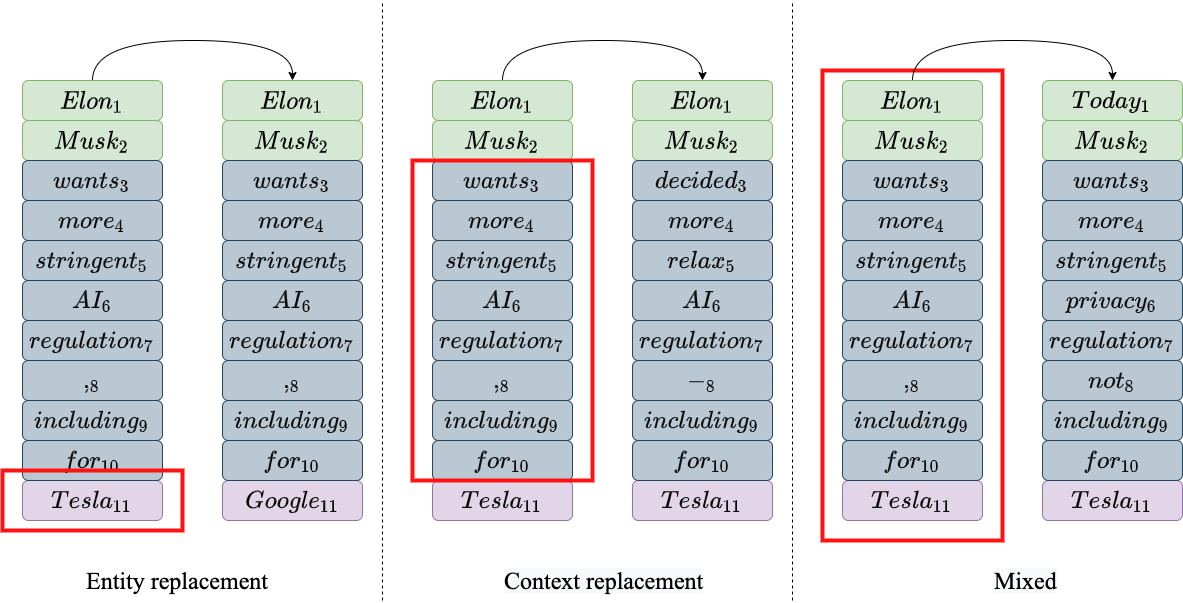}
    \caption{Sampling approaches example\footnotemark for the MLM data augmentation. Gray color refers to the tokens with the entity type ''O`` (context), green color refers to the \texttt{PER} entity type and purple color refers to the \texttt{ORG} entity type. Red square represents the subset of tokens which is used for replacement.}
    \label{fig:mlm-sampling}
\end{figure*}

Entity spans (entities of arbitrary length) make the training sentences used in NER tasks special. Since there is no guarantee that a predicted token belongs to the same entity type as an original token, it is important to ensure that the masked token is not in the middle of the entity span and that the existing label is not damaged. In this work, we propose three different types of token selection inside and outside of entity spans: 
\begin{itemize}
    \item \textbf{Entity replacement}: Collect entity spans of length one in the sentence and randomly select the entity span to be replaced. In this case, exactly one entity in the sentence is replaced. The sentences without entities or with longer entity spans are skipped.
    \item \textbf{Context replacement}:  We consider tokens with the label ``O'' as context and alternate between two setups: (1) Select only context tokens before and after entities, and (2) select a random subset of context tokens among all context tokens. 
    \item \textbf{Mixed}: Select uniformly at random the number of masked tokens between two and the sentence length among all tokens in the sentence.
\end{itemize}
The first approach allows only one entity to be generated and thus benefits from conditioning to the full sequence context. However, it does not guarantee the correct labeling for the generated token. The disadvantage of the second approach is that we do not generate new entity information, but only generate a new context for the existing entity spans. Even if a new entity type is generated, it has the original ``O'' label without a NER classification pipeline. The disadvantage of the third approach is that the token may be selected in the middle of the entity span and the label is no longer relevant. The sampling approaches depicted on the Figure~\ref{fig:mlm-sampling}. In addition, the number of replaced tokens should be properly tuned to avoid inadequate generation. In this work, we do not set any boundaries for maximum token replacement and leave such investigation to future work.

\footnotetext{Given example is taken from \url{https://artificialintelligence-news.com}}

\subsection{Order of Generation}
In our method, we predict exactly one mask token per time. Our sampling approaches allow multiple tokens to be replaced. Therefore we have two possible options for the generation order: 
\begin{itemize}
    \item \textbf{Independent}: Each consecutive masking and prediction is made on top of the original sequence. 
    \item \textbf{Conditional}: Each consecutive masking and prediction is made on top of the prediction of the previous step. 
\end{itemize}

\subsection{Criterion}
The criterion is an important part of the generation process. On the one hand, we want our synthetic sequence to be reliable (highest token probability), on the other hand, it should differ as much as possible from the original sequence (high distance). We propose two criteria for choosing the best token from the five-best predictions: 
\begin{itemize}
    \item \textbf{Highest probability (top token)}: Choose the target token only based on the MLM probability for that token.
    \item \textbf{Highest probability and distance (joint criterion)}: Choose the target token based on the product of the MLM probability for the token and Levenshtein distance \cite{L96} between the original sentence and the sentence with the new token.
\end{itemize}

Regardless of the combination of the parameters, the sentences must be changed. As a result, we guarantee that there is no duplication in our synthetic data with the original dataset. 

\subsection{Discussion}
The main disadvantage of using a language model (LM) for the augmentation of NER datasets is that the LM does not take into account the labeling of the sequence and the prediction of the masked token, which only depends on the surrounding tokens. As a result, we lose important information for decision-making. Incorporating label information as described in \newcite{WuLZHH19} into the MLM would be the way to tackle this problem. 

Another way to reduce the noise in the generated dataset is to apply a filtering step to the generation pipeline. One way to incorporate filtering into the augmentation process is to set the threshold for the MLM token probabilities: If the probability of the predicted token is less than a threshold, we ignore such prediction. However, the problem of misaligning token labels is not resolved. Therefore, we adapt our proposed confidence measure from Section \ref{sec:approach-self-training} for filtering.

In this work, we do not discuss the selection of the MLM itself as well as the effects of fine-tuning on the specific task.

\section{Experiments}
\subsection{Datasets}
We test our data adaptation approaches with three different NER datasets: CoNLL \cite{SD03}, W-NUT \cite{D17} and GermEval \cite{B14}.

All datasets have the original labeling scheme as \texttt{BIO}, but following \citet{LB16} we convert it to the \texttt{IOBES} scheme for training and evaluation. 
For our baseline models, we do not use any additional data apart from the provided training data. Development data is only used for validation. For CoNLL we skip all document boundaries. The statistics for the datasets are shown in Table \ref{tab:dataset-sizes}.\footnote{Further details on the used datasets can be found in Appendix \ref{app:dd}}

\begin{table}[ht]
\centering
\begin{tabular}{|l|l|r|r|r|}
\hline
\textbf{Dataset} & {\textbf{train}} & {\textbf{dev}} &   {\textbf{test}}                                               \\ \hline
CoNLL       & 14041     & 3250  & 3453\\ \hline
W-NUT       & 3394      & 1008  & 1287 \\ \hline
GermEval    & 24001     & 2199  & 5099 \\ \hline

\end{tabular}
\caption{Dataset sizes in number of sentences.}
\label{tab:dataset-sizes}
\end{table}

\subsection{Model Description}
\label{sec:model}
The Bidirectional LSTM - Conditional Random Field (BiLSTM-CRF) model \cite{LB16} is a widely used architecture for NER tasks. Together with  pre-trained word embeddings, it surpasses other neural architectures. We use the BiLSTM-CRF model implemented in the \textit{Flair}\footnote{\url{https://github.com/zalandoresearch/flair/}} framework version 0.5, which delivers the state-of-the-art performance.

The BiLSTM-CRF model consists of 1 hidden layer with 256 hidden states. Following \newcite{RG17}, we set the initial learning rate to 0.1 and the mini-batch size to 32. For each task, we select the best performing embedding from all embedding types in \textit{Flair}.
For training models with CoNLL data, we use pre-trained \textit{GloVE} \citep{PS14} word embedding \citep{GB18} together with the Flair embedding \citep{AB18} as input into the model.
For W-NUT experiments, we use \textit{roberta-large} embedding provided by \textit{Transformers} library \cite{WH19}. 
German \textit{dbmdz/bert-base-german-cased} embedding is used for experiments with the GermEval dataset.

\subsection{Unlabeled Data}
Additional unlabeled data is required for self-training.
To match the domain of the test data, we collect the data from the sources mentioned in the individual task descriptions.

\paragraph{W-NUT}
Like the test data, the data for W-NUT consists of user comments from Reddit, which were created in April 2017\footnote{\url{https://files.pushshift.io/reddit/comments/}} (comments in the test data were created from January to March 2017), as well as titles, posts and comments from StackExchange, which were created from July to December 2017\footnote{\url{https://archive.org/download/stackexchange}} (the content of the test data was created from January to May 2017).
The documents are filtered according to length and community as described in the task description paper and tokenized with the \textit{TweetTokenizer} from \textit{nltk}\footnote{\url{https://www.nltk.org/api/nltk.tokenize.html}}.

\paragraph{CoNLL}
The data was sampled from news articles in the Reuters corpus from October and November 1996.
The sentences are tokenized using \textit{spaCy}\footnote{\label{note:spacy}\url{https://github.com/explosion/spaCy}} and filtered (by removing common patterns like the date of the article, sentences that do not contain words and sentences with more than 512 characters as this is the length of the longest sentence in the CoNLL training data).

\paragraph{GermEval}
We randomly sampled additional data from sentences extracted from news and Wikipedia articles provided by the Leipzig Corpora Collection\footnote{\url{https://wortschatz.uni-leipzig.de/de/download}}.
In addition to tokenizing the sentences using spaCy, we do not do any additional preprocessing or filtering.

\subsection{Self-training}
\label{sec:experiments-self-training}
Before applying the approach described in \Cref{sec:approach-self-training}, we need to find the thresholds \(t\) for the confidence measures \(c_1\) and \(c_2\) for each corpus.
We evaluate both confidence measures on the development sets of the three corpora.
One way to evaluate confidence measures is to calculate the confidence error rate (CER).
It is defined as the number of misassigned labels (i.e. confidence is above the threshold and the prediction of the model is incorrect or the confidence is below the threshold and the prediction is correct) divided by the total number of samples.

\begin{figure}
    \centering
    \begin{center}
    \input{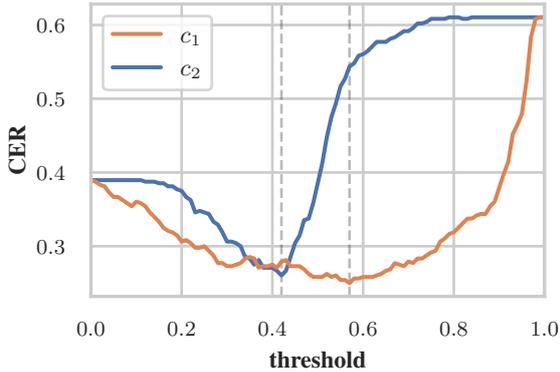}
    \end{center}
    \caption{CERs for \(c_1\) (orange) and \(c_2\) (blue) with different threshold values on the W-NUT development set. Vertical dashed lines represent $\hat t_1$ and $\hat{t_2}$.}
    \label{fig:cer_per_threshold}
\end{figure}

\Cref{fig:cer_per_threshold} shows the CER of \(c_1\) and \(c_2\) on the development set of W-NUT for different threshold values \(t\).
For the threshold of $0.0$ or $1.0$ the CER degrades to the percentage of incorrect or correct predictions as either all or no confidence values are above the threshold.
For \(c_2\) there is a clear optimum at \(\hat t_2 = 0.42\) and for larger and smaller thresholds the CER rises rapidly.

In contrast, the optimum for \(c_1\) at \(\hat t_1 = 0.57\) is not as pronounced.
This motivated us not only to choose the best value in terms of CER, but also a lower threshold \(t'_1 = 0.42\) with slightly worse CER.
In this way, we include more sentences where the model is less confident without introducing too many additional errors.
The threshold values for CoNLL and GermEval are selected analogously.
\Cref{tab:confidence_thresholds} provides an overview of all threshold values that are used in all subsequent experiments.

\begin{table}
\begin{tabular}{|l|r|r|r|}
\hline
                         & \textbf{W-NUT} & \textbf{CoNLL} & \textbf{GermEval} \\ \hline
$\hat t_1$ & 0.57           & 0.83           & 0.63              \\ \hline
$t'_1    $ & 0.42           & 0.70           & 0.50              \\ \hline
$\hat t_2$ & 0.42           & 0.50           & 0.47              \\ \hline
\end{tabular}%
\centering
\caption{Selected confidence threshold values.}
\label{tab:confidence_thresholds}
\end{table}



The unlabeled data is annotated using the baseline models described in \Cref{sec:approach-self-training} (we choose the best runs based on the score on the development set) and is filtered based on the different confidence thresholds.
Then we sample a random subset of size \(k\) from these remaining sentences.
For tasks where the data comes from different sources, e.g. news and Wikipedia for GermEval, we uniformly sample from the different sources to avoid that a particular domain is overrepresented.
The selected additional sentences are then appended to the original set of training sentences to create a new training set that is used to retrain the model from scratch.

\begin{figure}
    \centering
    \begin{center}
    \input{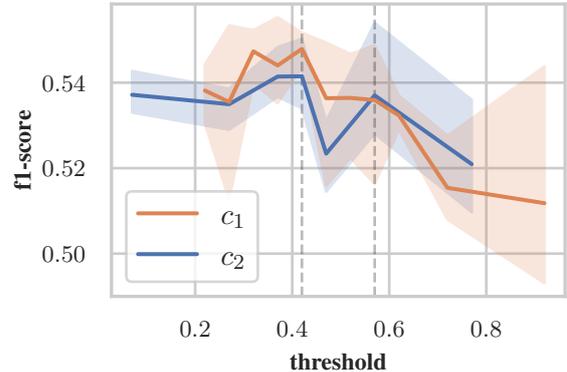}
    \end{center}
    \caption{Average F1 scores and standard deviation (shaded area) of 3 runs on the test set of W-NUT after retraining the model on additional data selected using different confidence measures (color) and thresholds.}
    \label{fig:wnut_different_thresholds}
\end{figure}

To validate our selection strategy, we test our pipeline with different confidence thresholds for both confidence measures.
\Cref{fig:wnut_different_thresholds} shows the results on the test set of W-NUT.
For each threshold, 3394 sentences are sampled, i.e. the size of the training set is doubled.
The results confirm our selection strategy.
\(t'_1\) and \(\hat t_2\) give the best results of all tested threshold values.
In particular, \(t'_1\) performs better than \(\hat t_1\).

\begin{table*}[ht] 
\begin{tabular}{|l|l|r|r|r|r|r|r|} 
\hline
 \multicolumn{2}{|c|}{} & \multicolumn{2}{c|}{\textbf{W-NUT}} & \multicolumn{2}{c|}{\textbf{CoNLL}} & \multicolumn{2}{c|}{\textbf{GermEval}}   \\ 
 \cline{3-8}
 \multicolumn{2}{|c|}{} & \textbf{$\Delta$ sen.} & \textbf{F1} & \textbf{$\Delta$ sen.} & \textbf{F1} & \textbf{$\Delta$ sen.} & \textbf{F1} \\ \hline
 1  & baseline            & $+0\% $      & $52.7 \pm 2.48$ & $ +0\% $     & $92.6 \pm 0.18$ & $ +0\% $     & $ 86.3 \pm 0.06 $ \\ \hline
 2  & $c_1 \geq \hat t_1$ & $+50\% $ & $54.2 \pm 0.35$ & $ +50\% $  & $92.5 \pm 0.06$ & $ +50\% $ & $ 86.0 \pm 0.08 $ \\
 3  & $c_1 \geq \hat t_1$ & $+100\% $ & $53.6 \pm 1.41$ & $ +100\% $ & $92.5 \pm 0.12$ & $ +100\% $ & $ 86.1 \pm 0.26 $ \\
 4  & $c_1 \geq \hat t_1$ & $+200\% $ & $53.5 \pm 0.53$ & $ +200\% $ & $92.4 \pm 0.08$ & $ +200\% $ & $ 86.3 \pm 0.14 $ \\ \hline
 5  & $c_1 \geq t'_1$     & $+50\% $ & $53.7 \pm 1.95$ & $ +50\% $  & $92.5 \pm 0.02$ & $ +50\% $ & $ 86.1 \pm 0.21 $ \\
 6  & $c_1 \geq t'_1$     & $+100\% $ & \mathcellbold{$54.8 \pm 0.33$} & $ +100\% $ & $92.6 \pm 0.09$ & $ +100\% $ & $ 86.2 \pm 0.12 $ \\
 7  & $c_1 \geq t'_1$     & $+200\% $ & $53.5 \pm 0.29$ & $ +200\% $ & $92.5 \pm 0.06$ & $ +200\% $ & \mathcellbold{$ 86.4 \pm 0.03 $} \\ \hline
 8  & $c_2 \geq \hat t_2$ & $+50\% $ & $54.6 \pm 0.42$ & $ +50\% $  & \mathcellbold{$92.7 \pm 0.04$} & $ +50\% $ & $ 86.0 \pm 0.16 $ \\
 9  & $c_2 \geq \hat t_2$ & $+100\% $ & $54.2 \pm 0.98$ & $ +100\% $ & $92.6 \pm 0.06$ & $ +100\% $ & \mathcellbold{$ 86.4 \pm 0.15 $} \\
 10 & $c_2 \geq \hat t_2$ & $+200\% $ & $54.5 \pm 0.43$ & $ +200\% $ & \mathcellbold{$92.7 \pm 0.02$} & $ +200\% $ & $ 86.3 \pm 0.05 $ \\ \hline
\end{tabular}
\centering
\caption{Results of self-training.}
\label{tab:self-training}
\end{table*}

\Cref{tab:self-training} shows the results of self-training on all three datasets.
For each of them, we test the three selection strategies by sampling new sentences in the size of 0.5 times, 1 times and 2 times the size of the original training data.
For W-NUT we get up to 2\% of the absolute improvements in the F1 score over the baseline.
On larger datasets like CoNLL and GermEval these effects disappear and we only get improvements of up to 0.1\% and in some cases even deterioration.

\subsection{MLM-based Data Augmentation}
We follow the approach explained in Section \ref{section:mlm} and generate synthetic data using pre-trained models from the Transformers library. We concatenate original and synthetic data and train the NER model on the new dataset. We test all possible combinations of the augmentation parameters from Section \ref{section:mlm} on the W-NUT dataset. Table \ref{tab:wnut-params-search} shows the result of the augmentation. When sampling with one entity, there is no difference between independent and conditional generation, since only one token in a sentence is masked. We therefore only carry out an independent generation for this type of sampling. We report an average result among 3 runs along with a standard deviation of the model with different random seeds.

\begin{table*}[ht]
\centering
\begin{tabular}{|c|c|c|c|c|r|r|}
\hline
                               & & \textbf{sampling} & \textbf{generation}  & \textbf{criterion} & \textbf{$\Delta$ sen.} & \textbf{F1} \\ \hline
1 & baseline                       & -          & -                           & -               & $+0.0\%$ & $52.7 \pm 2.48$        \\ \hline
2&                                & \multirow{2}{*}{entity}	& \multirow{2}{*}{independent}	& top token & $ +24.4\% $ & $53.7 \pm 0.91$ \\ \cline{5-7} 
 3 &                              &  & & joint & $ +24.7\% $ & $54.6 \pm 0.50$ \\ \cline{3-7} 
 4 &                               & \multirow{4}{*}{mixed} & \multirow{2}{*}{conditional}                        & top token             & $ +98.7\% $ & $52.3 \pm 1.25$       \\ \cline{5-7} 
 5 &                              &      &                     & joint             & $ +99.7\% $ & $51.7 \pm 1.36$      \\ \cline{4-7} 
 6 &                              &     & \multirow{2}{*}{independent}                   & top token         & $ +98.6\% $ & $53.7 \pm 0.89$       \\ \cline{5-7} 
 7 &                              &        &                     & joint         & $ +99.7\% $ & $53.3 \pm 0.61$       \\ \cline{3-7} 
 8 &                              & \multirow{4}{*}{context}        & \multirow{2}{*}{conditional}                        & top token             & $ +33.8\% $ & \mathcellbold{$56.3 \pm 1.21$}       \\ \cline{5-7} 
9 &                               &    &                        & joint         & $ +35.8\% $ & \mathcellbold{$ 55.6 \pm 1.12 $}       \\ \cline{4-7} 
10 &                               &        & \multirow{2}{*}{independent}                     & top token             & $ +33.8\% $ & \mathcellbold{$ 55.0 \pm 1.16 $}      \\ \cline{5-7} 
11 &                               &       &                         & joint         & $ +35.8\% $ & \mathcellbold{$ 56.0 \pm 0.06 $}      \\ \cline{3-7} 
12 &                               & \multirow{4}{*}{random context}    & \multirow{2}{*}{conditional}                        & top token             & $ +96.8\% $ & \mathcellbold{$ 54.9 \pm  0.40 $}       \\ \cline{5-7} 
13 &                               &   &                       & joint         & $ +99.7\% $ & $54.5 \pm 1.21$      \\ \cline{4-7} 
14 &                               &   & \multirow{2}{*}{independent}                      & top token             & $ +96.9\% $ & $53.7 \pm 0.93$       \\ \cline{5-7} 
15 & \multirow{-14}{*}{MLM DA} &    &                      & joint         & $ +99.7\% $ & $53.5\pm 2.40$       \\ \hline
\end{tabular}
\caption{Results of the MLM-based augmentation on the W-NUT dataset. \texttt{entity} refers to the sampling tokens from entity spans of length one, \texttt{mixed} means sampling from the complete sequence, \texttt{context} indicates sampling from the entity span context, \texttt{random context} denotes sampling from random context labels. \texttt{conditional} refers to the conditional generation and \texttt{independent} refers to the independent generation type. The \texttt{top token} criterion selects the token based on the highest probability, and the \texttt{joint} criterion takes into account the token probability and the Levenshtein distance.}
\label{tab:wnut-params-search}
\end{table*}

W-NUT and CoNLL datasets are augmented using a pre-trained English BERT model\footnote{\url{https://huggingface.co/bert-large-cased-whole-word-masking}} and GermEval with a pre-trained German BERT model\footnote{\url{https://huggingface.co/bert-base-german-cased}} respectively. We do not fine-tune these models.

Sampling from the context of the entity spans shows significant improvements on W-NUT test set. First of all, it includes implicit filtering: Only the sentences with the entities are selected and replaced. Therefore, compared to other methods, we add less new sentences (except when replacing entities). Second of all, since replacing tokens with a language model should result in the substitution with similar words, the label is less likely to be destroyed while context tokens are replaced.

On the other hand, the mixed sampling strategy performs the worst among all methods. We believe that this is the effect when additional noise is included in the dataset (by noise we mean all types of noise, e.g. incorrect labeling, grammatical errors, etc). Allowing masking of words up to sequence in some cases destroys the sentence, e.g. incorrect and multiple occurrences of the same words can occur. In Appendix~\ref{app:mlm-app} we present the examples of augmented sentences for each augmentation approach and each dataset. Additionally, we report the average number of masked token.

\begin{figure}[!t]
    \centering
    \input{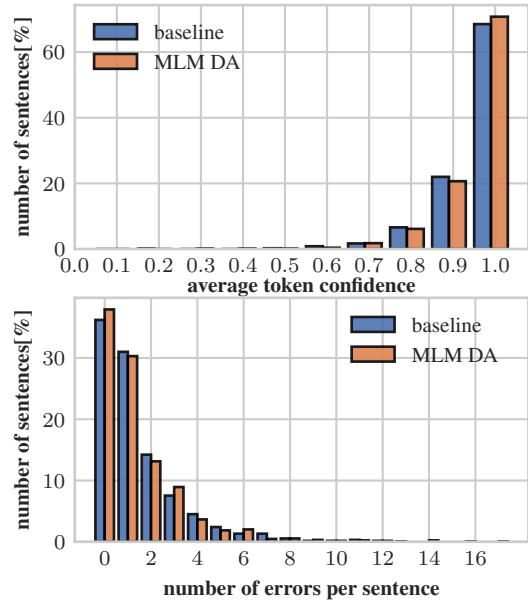}
    \caption{Average confidence score and the error per sentence on W-NUT test data. \texttt{MLM DA} refers to the setup of line 8 in Table~\ref{tab:wnut-params-search}}
    \label{fig:conf}
\end{figure}

To analyze the resulting models, we plot the average confidence scores of the test set as well as the number of errors per sentence for the best baseline model and best augmented model. We use the best baseline system with 54.6\% F1 score and the best model corresponding to the setup of line 8 in Table~\ref{tab:wnut-params-search} with 57.4\% F1 score. We count the error every time the model predicts a correct label with low confidence or an incorrect label with high confidence. We set high and low confidence to be 0.6 and 0.4 respectively. Figure \ref{fig:conf} shows that the augmented model makes a more reliable prediction than the best baseline system model.

We repeat the promising MLM generation pipeline on the CoNLL and GermEval datasets. These datasets contain more entities in the original data. In addition, even though the entity replacement sampling did not work well on W-NUT dataset, we repeat these experiments, since generating new entities is the most interesting scenario for using the MLM augmentation.

Although the MLM-based data augmentation leads to improvements of up to 3.6\% F1 score on the W-NUT dataset, Table \ref{tab:conll-germeval-mlm} shows that such effect disappears when we apply our method to larger and cleaner datasets such as CoNLL and GermEval. We believe there are several reasons for that. First, our MLM-based data augmentation method does not guarantee the accuracy of the labeling after augmentation. So for larger datasets, there are many more possibilities to increase the noise of the corpus. Moreover, we do not study how well pre-trained models suit the specific task, which might be crucial\textbf{} for the DA. Besides, for GermEval augmentation, we use the BERT model with three times fewer parameters than for W-NUT and CoNLL.

\begin{table*}[ht]
\centering
\resizebox{\linewidth}{!}{%
\begin{tabular}{|c|c|c|c|c|r|r|r|r|r|}
\hline
\multicolumn{5}{|c|}{} & \multicolumn{2}{c|}{\textbf{CoNLL}} & \multicolumn{2}{c|}{\textbf{GermEval}} \\ \hline
                             \multicolumn{2}{|c|}{}  & \textbf{sampling} & \textbf{generation}  & \textbf{criterion} & \textbf{$\Delta$ sen.} & \textbf{F1} & \textbf{$\Delta$ sen.} & \textbf{F1}\\ \hline
1& baseline                       & -          & -                           & -               & $ +0.0\% $    & \mathcellbold{$92.6 \pm 0.18$}       & 0.0\% & \mathcellbold{$86.3 \pm 0.06$}\\ \hline
3& & entity & independent & joint & $ +57.9\% $ &  $91.5 \pm 0.10$ & $ +47.9\% $ & $85.9 \pm 0.06$ \\ \cline{3-9}
8&                               &    \multirow{4}{*}{context}     & \multirow{2}{*}{conditional}                        & top token             &   $ +65.7\% $   & $92.4 \pm 0.12$        & $ +51.4\% $ & $86.1 \pm 0.26$ \\ \cline{5-9} 
9&                               &         &                          & joint         &   $ +72.2\% $   & $92.3 \pm 0.06$      & $ +58.5\% $ & $86.0 \pm 0.15$ \\ \cline{4-9} 
10 &                               &        & \multirow{2}{*}{independent}                          & top token         &   $ +65.7\% $   &    $92.5 \pm 0.06$   & $ +51.4\% $ & $86.1 \pm 0.15$ \\ \cline{5-9} 
11 &                               &        &                         & joint         &   $ +72.2\% $   & $92.2 \pm 0.17$       & $ +58.5\% $  & $86.0 \pm 0.20$ \\ \cline{3-9} 
12 & \multirow{-6}{*}{MLM DA}  & rand. cont. & conditional                        & top token         &  $ +85.1\% $    & $92.1 \pm 0.15$     & $ +94.1\% $ & $86.1 \pm 0.10$ \\
  \hline
\end{tabular}
}
\caption{Results of the MLM-based data augmentation on CoNLL and GermEval datasets. The row numbers refer to the row numbers of the Table~\ref{tab:wnut-params-search}.}
\label{tab:conll-germeval-mlm}
\end{table*}
 
 \subsubsection{Filtering of Augmented Data}
 As discussed in Section \ref{section:mlm}, an additional data filtering step can be applied on top of the augmentation process. We report results on two different filtering methods: First, we set a threshold for the probability of the predicted token (in our experiments we use the probability 0.5); Second, we filter sentences by minimum confidence scores as discussed in Section \ref{sec:approach-self-training}. We set the minimum confidence score according to Table \ref{tab:confidence_thresholds}. We apply filtering to the worst and best-performing model according to the numbers in Table~\ref{tab:wnut-params-search}. The filtering results on W-NUT are shown in Table \ref{tab:wnut-filtering}.

\begin{table}[ht]
    \centering
    \resizebox{0.75\linewidth}{!}{%
    \begin{tabular}{|c|r|c|r|c|}
    \hline
        & \textbf{$\Delta$ sen.} & \textbf{filtering} & \textbf{F1} \\ \hline
         & $+99.7\%$ & - & $51.7 \pm 1.36$ \\ \cline{2-4}
         & $+86.3\%$ & token prob. &\mathcellbold{$54.3 \pm 0.31$}   \\ \cline{2-4}
         \multirow{-3}{*}{5} & $+59.5\%$  & min. conf.&  $51.2 \pm 0.60$ \\ \hline
         
         & $+33.8\%$ & - & \mathcellbold{$56.3 \pm 1.21$} \\ \cline{2-4}
         & $+13.8\%$ & token prob. & $53.3 \pm 2.00$  \\ \cline{2-4}
         \multirow{-3}{*}{9} & $+10.4\%$  & min. conf. &  $51.7 \pm 2.10$ \\ \hline

    \end{tabular}
    }
    \caption{F1 scores of using filtered augmented data on W-NUT. The row numbers refer to the row numbers of the Table~\ref{tab:wnut-params-search}.}
    \label{tab:wnut-filtering}
\end{table}

 In the case of the worst model, filtering based on the token probability improve the performance of the model by 2.6\% compared to the unfiltered one. Filtering by confidence score does not improve the performance, but significantly reduces the standard deviation of the score. The results are expected, since by using token probability we increase the sentence reliability and completely change the synthetic data, while using the confidence score we filter on the same synthetic data. In the case of the better model, we see the opposite trend. Here filtering leads to performance degradation and an increase in the standard deviation.
 
 We apply the same filtering techniques for CoNLL and GermEval. Table \ref{tab:conll-geval-filtering} shows the results for 3 different models. We choose the best, the worst and the model with the highest number of additional sentences for filtering. In the case of the worst model, the performance is improved by 1.1\% F1 score with the minimum confidence filtering for CoNLL and 0.5\% F1 score for GermEval compared to the unfiltered version. However, for the best model, the results remain at the same level and the baseline systems are not improved.
 
\begin{table}[ht]
    \centering
    \resizebox{\linewidth}{!}{%
    \begin{tabular}{|c|c|r|r|r|r|}
    \hline
         &  & \multicolumn{2}{c|}{\textbf{CoNLL}}&\multicolumn{2}{c|}{\textbf{GermEval}} \\ \cline{3-6}
         & \textbf{filtering} & \textbf{$\Delta$ sen.}  & \textbf{F1} & \textbf{$\Delta$ sen.}  & \textbf{F1}\\ \hline
         & none & $+57.9\%$ & $91.5 \pm 0.10$ & $+47.9\%$ & $85.9 \pm 0.06$ \\ \cline{2-6}
         & tok. prob. & $+7.8\%$ & $92.4 \pm 0.15$ & $+13.1\%$  & $86.1 \pm 0.29$\\ \cline{2-6}
         \multirow{-3}{*}{3} & min. conf. & $+13.5\%$ & $92.6 \pm 0.15$  &  $+13.9\%$ & \mathcellbold{$ 86.4 \pm 0.12 $}\\ \hline
         
         & none & $+65.7\%$   &    $92.5 \pm 0.06$  & $+51.5\%$ & $86.1 \pm 0.15$ \\ \cline{2-6}
         &tok. prob. & $+22.5\%$ & $92.5 \pm 0.15$& $+34.5\%$ & \mathcellbold{$ 86.3 \pm 0.21 $} \\ \cline{2-6}
         \multirow{-3}{*}{10} & min. conf. & $+52.1\%$ & \mathcellbold{$ 92.6 \pm 0.20 $} & $+23.9\%$ & $86.1 \pm 0.10$\\ \hline
         
         & none & $+85.1\%$   & $92.1 \pm 0.15$  & $+94.1\%$ & $86.1 \pm 0.10$ \\ \cline{2-6}
         &tok. prob. & $+42.5\%$ & \mathcellbold{$ 92.8 \pm 0.06 $} & $+76.1\%$ & $86.1 \pm 0.00$\\ \cline{2-6}
         \multirow{-3}{*}{12} & min. conf. & $+58.9\%$ & \mathcellbold{$ 92.6 \pm 0.12 $} &  $+62.3\%$ & $86.0 \pm 0.21$\\ \hline

    \end{tabular}%
    }
    \caption{F1 scores of using filtered augmented data on CoNLL and GermEval. The first line represents the augmentation method from Table \ref{tab:wnut-params-search}.}
    \label{tab:conll-geval-filtering}
\end{table}
 
  Although we do not achieve significant improvements compared to the baseline system, we see a potential in the MLM-based augmentation with the combination with filtering.
  
  \section{Discussion and Future Work}
In this work, we present results of data adaptation methods on various NER tasks. We show that MLM-based data augmentation and self-training approaches lead to improvements on the small and noisy W-NUT dataset. 

We propose two different confidence measures for self-training and empirically estimate the best thresholds. Our results on the W-NUT dataset show the effectiveness of the selection strategies based on those confidence measures. 

For MLM-based data augmentation, we suggest multiple ways of generating synthetic NER data. Our results show that even without generating new entity spans we are able to achieve better results.

For future work, we would like to incorporate label information into the augmentation pipeline by either conditioning the token predictions on labels or adding additional classification steps on top of the token prediction. Another important question is the choice of the MLM and the impact of task-specific fine-tuning. Further investigations into the filtering step should also be carried out.

For both self-training and MLM-based data augmentation we would like to improve the integration in the training process.
The contribution of the original training data to the loss function could be increased or additional data could be weighted by their confidence.
Finally, we would like to test whether we can combine the two methods to achieve additional improvements.

\section*{Acknowledgements}
This work has received funding from the European Research Council (ERC) under the European Union's Horizon 2020 research and innovation programme (grant agreement No 694537, project ``SEQCLAS''). The work reflects only the authors' views and the European Research Council Executive Agency (ERCEA) is not responsible for any use that may be made of the information it contains.

\bibliographystyle{acl_natbib}
\bibliography{anthology,acl2021}

\appendix
\clearpage
\section{Data Description}
\label{app:dd}
In our work we use three NER datasets:
\begin{itemize}
    \item CoNLL 2003 \cite{SD03} contains news articles from the Reuters\footnote{\url{https://trec.nist.gov/data/reuters/reuters.html}} corpus. The annotation contains 4 entity types \texttt{person}, \texttt{location}, \texttt{organization}, \texttt{miscellaneous}. We remove the document boundary information for our experiments. 
    \item W-NUT 2017 \cite{D17} contains texts from Twitter (training data), YouTube (development data), StackExchange and Reddit (test data). 
    The annotation contains 6 entity types: \texttt{person}, \texttt{location}, \texttt{corporation}, \texttt{product}, \texttt{creative-work},  \texttt{group}
    \item GermEval 2014 \cite{B14}: contains the data from the German Wikipedia and news Corpora. The annotation contains 12 entity types: \texttt{location}, \texttt{organization}, \texttt{person}, \texttt{other}, \texttt{location deriv}, \texttt{location part}, \texttt{organization deriv}, \texttt{organization part}, \texttt{person deriv}, \texttt{person part}, \texttt{other deriv},  \texttt{other part}.  
\end{itemize}
Table~\ref{tab:dataset-sizes-det} shows detailed statistics of those datasets. Together with number of entities, tokens and sentences we report the percentage of the labelled tokens among all the tokens.
\begin{table}[ht]
\centering
\resizebox{0.8\linewidth}{!}{%
\begin{tabular}{|l|l|r|r|r|}
\hline
\textbf{Dataset} & &{\textbf{train}}                                             & {\textbf{dev}}                                                 & {\textbf{test}}                                               \\ \hline
    & \#sentences & 14041 & 3250 & 3453\\ 
    & \#entities & 23500 & 5943 & 5649\\
    & \#tokens & 203621 & 51362& 46435\\
& \#entity types & 4 & 4 & 4 \\
\multirow{-4}{*}{CoNLL} & \%labelled & 16.7 & 16.8 & 17.5 \\    \hline

    &   \#sentences & 3394 & 1008  & 1287 \\ 
    &   \#entities & 1976 & 836 & 1080 \\
    & \#tokens & 62730 & 15723 & 23394 \\
& \#entity types & 6 & 6 & 6 \\
\multirow{-4}{*}{W-NUT} & \%labelled & 5.0 & 7.9 & 7.4 \\   \hline

& \#sentences & 24001 & 2199 & 5099\\ 
& \#entities & 29077 & 2674 & 6178 \\
& \#tokens & 452790 & 41635 & 96475 \\
& \#entity types & 12 & 12 & 12 \\
\multirow{-4}{*}{GermEval}  & \%labelled & 9.3 & 9.5 & 9.3\\   \hline
\end{tabular}%
}
\caption{Dataset sizes in number of sentences, tokens and entities. Here, entity means the entity span, e.g. \texttt{European Union} is considered as one entity.}
\label{tab:dataset-sizes-det}
\end{table}

\section{MLM-based Data Augmentation}
\label{app:mlm-app}
\subsection{Data statistics}
The number of masked tokens solely depends on the augmentation strategy discussed in section~\ref{section:mlm}.  Table~\ref{tab:wnut-masked-avg} reports the average number of masked tokens in the sentence on W-NUT dataset for each augmentation strategy. Table~\ref{tab:conll-masked-avg}
and Table~\ref{tab:germeval-masked-avg} show the average number of masked tokens in the sentence for the most promising augmentation strategies for CoNLL and GermEval tasks.

\begin{table}[ht]
\centering
\resizebox{.45\textwidth}{!}{%
\begin{tabular}{|c|c|c|c|c|}
\hline
     \textbf{sampling} & \textbf{generation}  & \textbf{criterion} & \textbf{$\Delta$ sen.} & \textbf{Masked} \\ \hline                            
    \multirow{2}{*}{entity}	& \multirow{2}{*}{independent}	
    & top token & $ +24.4\% $ &  1.2 \\ \cline{3-5} 
    & & joint & $ +24.7\% $ &  1.2 \\ \hline 
    
     \multirow{4}{*}{mixed} & \multirow{2}{*}{conditional}  
     & top token & $ +98.7\% $ & 7.4 \\  \cline{3-5}
     & & joint & $ +99.7\% $ & 8.8 \\ \cline{2-5} 
     & \multirow{2}{*}{independent}    
     & top token         & $ +98.6\% $ &    7.0    \\ \cline{3-5}
     & & joint& $ +99.7\% $ & 8.8 \\ \hline
    
     \multirow{4}{*}{context} & \multirow{2}{*}{conditional}                        
    & top token             & $ +33.8\% $ &  4.4      \\  \cline{3-5}
     & & joint & $ +35.8\% $ & 4.5\\ \cline{2-5} 
     & \multirow{2}{*}{independent}                     
    & top token             & $ +33.8\% $ & 4.3 \\ \cline{3-5} 
     & & joint & $ +35.8\%$  &     4.5 \\ \hline
    
     \multirow{4}{*}{random context} & \multirow{2}{*}{conditional}                        
    & top token             & $ +96.8\% $ &   7.1    \\  \cline{3-5}
     & & joint & $ +99.7\% $ & 8.1 \\ \cline{2-5}
     & \multirow{2}{*}{independent} & top token & $ +96.9\% $ & 6.9\\ \cline{3-5}
    &  & joint & $ +99.7\% $ & 8.1\\ \hline
\end{tabular}%
}
\caption{Average number of masked tokens for each augmentation strategy on W-NUT dataset. }
\label{tab:wnut-masked-avg}
\end{table}

\begin{table}[ht]
\centering
\resizebox{.45\textwidth}{!}{%
\begin{tabular}{|c|c|c|c|c|}
\hline
     \textbf{sampling} & \textbf{generation}  & \textbf{criterion} & \textbf{$\Delta$ sen.} & \textbf{Masked} \\ \hline                            
    entity	& independent
    &  joint & $ +57.9\% $ &  1.1 \\ \hline 
    
     \multirow{4}{*}{context} & \multirow{2}{*}{conditional}                        
    & top token             & $ +65.7\% $  &  3.4      \\  \cline{3-5}
     & & joint & $ +72.2\% $ & 6.4\\ \cline{2-5} 
     & \multirow{2}{*}{independent}                     
    & top token             & $ +65.7\% $ & 3.4 \\ \cline{3-5} 
     & & joint &  $ +72.2\% $ &   6.4   \\ \hline
    
     random context & conditional                       
    & top token             & $ +85.1\% $ &   4.5   \\   \hline
\end{tabular}%
}
\caption{Average number of masked tokens on CoNLL dataset.}
\label{tab:conll-masked-avg}
\end{table}

\begin{table}[ht]
\centering
\resizebox{.45\textwidth}{!}{%
\begin{tabular}{|c|c|c|c|c|}
\hline
     \textbf{sampling} & \textbf{generation}  & \textbf{criterion} & \textbf{$\Delta$ sen.} & \textbf{Masked} \\ \hline                            
    entity	& independent & joint & $ +47.9\% $ &  1.0 \\ \hline

     \multirow{4}{*}{context} & \multirow{2}{*}{conditional}                        
    & top token             & $ +51.4\% $ &   4.4     \\  \cline{3-5}
     & & joint & $ +58.5\% $ & 5.7 \\ \cline{2-5} 
     & \multirow{2}{*}{independent}                     
    & top token             &  $ +51.4\% $ & 4.3 \\ \cline{3-5} 
     & & joint & $ +58.5\% $   &   5.3   \\ \hline
    
     random context & conditional                       
    & top token             & $ +94.1\% $ &  6.0 \\ \hline
\end{tabular}%
}
\caption{Average number of masked tokens on GermEval dataset. }
\label{tab:germeval-masked-avg}
\end{table}

\label{app:example}
\subsection{Data Examples}
We show the data examples on different dataset by varying one augmentation parameter while keeping others unchanged. Table~\ref{tab:wnut-full-examples} shows the examples on W-NUT dataset. In Table~\ref{tab:geval-full-examples} and Table~\ref{tab:conll-full-examples} we collect the examples for GermEval and CoNLL.

\begin{table*}[ht]
    \centering
    \begin{tabularx}{\linewidth}{|c|c|L|}
    \hline
         \textbf{Parameter} & \textbf{Value} & \textbf{Example}  \\ \hline
            & - & \texttt{RT @Quotealicious: Today, I saw a guy driving a <corporation>Pepsi</corporation> truck,  drinking a <product>Coke</product>. MLIA \#Quotealicious} \\
            & entity & \texttt{RT @Quotealicious: Today, I saw a guy driving a <corporation>Pepsi</corporation> truck, drinking a \textbf{<product>beer</product>} MLIA \#Quotealicious}\\
            & context & \texttt{RT @Quotealicious : Today, I saw a guy driving a <corporation>Pepsi</corporation> \textbf{car}, drinking a <product>Coke</product>. MLIA \#Quotealicious} \\
            & random context & \texttt{\textbf{m me}: Today, I saw a \textbf{man} driving a <corporation>Pepsi</corporation> truck, \textbf{buying} a <product>Coke</product>. MLIA \#Quotealicious} \\
         \multirow{-12}{*}{Sampling }& mixed & \texttt{\textbf{m} @Quotealicious \textbf{Earlier} Today, I saw a guy driving a <corporation>Pepsi</corporation> truck, drinking a <product>Coke</product>. MLIA \#Quotealicious}\\ \hline

         & - & \texttt{What is everyone watching this weekend? <group>Twins</group>? <group>Vikings</group>? anyone going to see <creativework>Friday Night Lights</creativework>?} \\
        & independent & \texttt{What is everyone watching this weekend? <group>Twins</group>? <group>Vikings</group>? anyone going to see <creativework>\textbf{the} Night Lights</creativework>?}\\
         \multirow{-6}{*}{Order}& conditional & \texttt{What is \textbf{he} doing this weekend with \textbf{<group>the</group> \#\#ing} <group>Vikings</group>? anyone going to install <creativework>Friday Night \textbf{lights}</creativework>?}\\ \hline
         
         & - & \texttt{<person>Oscar</person>'s new favorite pass time is running as fast as he can from one end of the house to another yelling BuhBYYYYYE} \\
        & top token & \texttt{\textbf{<person>Jack</person>}'s new favorite pass time is running as fast as he can from one end of the house to another yelling BuhBYYYYYE}\\
         \multirow{-6}{*}{Criterion}& joint & \texttt{\textbf{<person>Ben</person>}'s new favorite pass time is running as fast as he can from one end of the house to another yelling BuhBYYYYYE}\\ \hline
    \end{tabularx}
    \caption{Data examples of W-NUT augmentation.}
    \label{tab:wnut-full-examples}
\end{table*}

\begin{table*}[ht]
    \centering
    \begin{tabularx}{\linewidth}{|c|c|L|}
    \hline
         \textbf{Parameter} & \textbf{Value} & \textbf{Example}  \\ \hline
            & - & \texttt{Zu einer Gebietsveränderung kam es 1822, als das vorher selbständige <LOC>Champsigna</LOC> nach <LOC>Soucia</LOC> eingemeindet wurde.}\\
            & entity & \texttt{Zu einer Gebietsveränderung kam es 1822, als das vorher selbständige <LOC>Champsigna</LOC> nach \textbf{<LOC>Paris</LOC>} eingemeindet wurde.} \\ 
            & context & \texttt{Zu einer Gebietsveränderung kam es 1822, als das vorher selbständige <LOC>Champsigna</LOC> nach <LOC>Soucia</LOC> \textbf{verlegt} wurde.} \\
            & random context & \texttt{Zu einer Gebietsveränderung kam es 1822, als das \textbf{damals} selbständige <LOC>Champsigna</LOC> nach <LOC>Soucia</LOC> eingemeindet wurde.} \\
         \multirow{-12}{*}{Sampling }& mixed & \texttt{Zu einer \textbf{Eingemeindung} kam es 1822, als \textbf{die damals} selbständige \textbf{<LOC>Dorf</LOC>} nach \textbf{<LOC>Turin</LOC> verlegt} wurde.}\\ \hline

         & - & \texttt{Aus diesem Grund wurde er Anfang Januar auch nach nur wenigen Tagen aus dem Klinikum <LOC>Jena</LOC> in eine Reha-Einrichtung am <LOC>Bodensee</LOC> verlegt.} \\
        & independent & \texttt{\textbf{Zu} diesem Grund wurde er Anfang Januar \textbf{und} nach nur \textbf{zwei} Tagen aus dem Klinikum <LOC>Jena</LOC> in \textbf{die} Reha-Einrichtung am \textbf{<LOC> Boden </LOC>} verlegt.}\\
         \multirow{-5}{*}{Order}& conditional & \texttt{Aus diesem Grund \textbf{wo ich} Anfang Januar auch nach nur wenigen Tagen aus dem Klinikum <LOC>Jena</LOC> in \textbf{die} Reha-Einrichtung am <LOC>Bodensee</LOC> verlegt.}\\ \hline
         
         & - & \texttt{Mit ihm der gleichen Meinung sind <PER>Pyrrhon</PER> und <PER>Erillus</PER> von <LOC>Karthago</LOC>.} \\
        & top token & \texttt{Mit ihm der gleichen Meinung sind <PER>Pyrrhon</PER> und \textbf{<PER>Gregor</PER>} von <LOC>Karthago</LOC>.}\\
         \multirow{-5}{*}{Criterion}& joint & \texttt{Mit ihm der gleichen Meinung sind \textbf{<PER>Alexander</PER>} und <PER>Erillus</PER> von <LOC>Karthago</LOC>.}\\ \hline
    \end{tabularx}
    \caption{Data examples of GermEval augmentation.}
    \label{tab:geval-full-examples}
\end{table*}

\begin{table*}[ht]
    \centering
    \resizebox{.95\textwidth}{!}{%
    \begin{tabularx}{\linewidth}{|c|c|L|}
    \hline
         \textbf{Parameter} & \textbf{Value} & \textbf{Example}  \\ \hline
            & - & \texttt{<PER>Christopher Reeve</PER> -- <PER>Reeve</PER> was best known for playing the comic book hero <PER>Superman</PER> in four movies but his greatest heroics came in real life.} \\
            & entity & \texttt{<PER>Christopher Reeve</PER> -- <PER>Reeve</PER> was best known for playing the comic book hero \textbf{<PER>Batman</PER>} in four movies but his greatest heroics came in real life .}\\
            & context & \texttt{<PER>Christopher Reeve</PER> \textbf{The} <PER>Reeve</PER> \textbf{is} best known for playing the comic book superhero <PER>Superman</PER> in four movies but his greatest heroics came in real life.} \\
            & random context & \texttt{<PER>Christopher Reeve</PER> -- <PER>Reeve</PER> \textbf{popular} best known for \textbf{popular popular popular} book hero <PER>Superman</PER> in four movies but his popular heroics came in real \textbf{popular popular}} \\
         \multirow{-18}{*}{Sampling }& mixed & \texttt{<PER>Christopher Reeve</PER> \textbf{The <PER>He</PER> is} best known for playing the comic book superhero <PER>Superman</PER> in \textbf{the films} but his greatest heroics came in real life.}\\ \hline

         & - & \texttt{Four weeks ago <ORG>Stagecoach </ORG> said it had agreed the deal in principle, and it expected to pay 110 million stg-plus for the firm, with <ORG>Swebus</ORG>' current owner, the state railway company.} \\
         & independent & \texttt{Four \textbf{days} ago <ORG>it</ORG> said it had \textbf{made} the deal in principle, and it expected to \textbf{raise} 110 million \textbf{euros to} the \textbf{operation contract including} <ORG>Swebus</ORG> ' current \textbf{employer being} the state railway company.}\\
         \multirow{-7}{*}{Order}& conditional & \texttt{\textbf{Two years} ago <ORG>Stagecoach</ORG> said it had \textbf{made} the deal in principle, and \textbf{was} expected to pay 110 million \textbf{marks} for the \textbf{operation}, with <ORG>Swebus</ORG>'\textbf{s} owner, the \textbf{Swedish} railway company.}\\ \hline
         
         & - & \texttt{<ORG>ZDF</ORG> said <LOC> Germany </LOC> imported 47,600 sheep from <LOC> Britain </LOC> last year, nearly half of total imports.} \\
         & top token & \texttt{\textbf{<ORG>He</ORG>} said \textbf{<LOC> they </LOC>} imported \textbf{more goods} from \textbf{<LOC> Germany </LOC> that} year, nearly half of \textbf{all number}.}\\
         \multirow{-6}{*}{Criterion}& joint & \texttt{<ORG>ZDF</ORG> \textbf{this <LOC> this </LOC> this} 47,600 sheep \textbf{this <LOC> this </LOC> this} year \textbf{this} nearly half of \textbf{this} imports.}\\ \hline
    \end{tabularx}%
    }
    \caption{Data examples of CoNLL augmentation.}
    \label{tab:conll-full-examples}
\end{table*}

\end{document}